\documentclass{article}

\usepackage[dblblindworkshop,nonanonymous]{neurips_2026}
\workshoptitle{SLM-Agents: 1st Workshop on SLMs for Agentic Systems}

\usepackage[utf8]{inputenc}
\usepackage[T1]{fontenc}
\usepackage{hyperref}
\usepackage{url}
\usepackage{booktabs}
\usepackage{multirow}
\usepackage{amsfonts}
\usepackage{amsmath}
\usepackage{algorithm}
\usepackage{algpseudocode}
\usepackage{nicefrac}
\usepackage{microtype}
\usepackage{xcolor}
\usepackage{graphicx}
\graphicspath{{figure/}}
\usepackage{enumitem}
\setlist[enumerate]{leftmargin=*,nosep,topsep=2pt,partopsep=0pt}

\title{Making Prospective Memory SLM-Shaped:\\
Typed Intention Stores for Small-Model Agents}

\author{%
  Jinqing Zhao \\
  Peking University \\
  \And
  Chengcan Wu \\
  Peking University \\
  \texttt{wuchengcan@stu.pku.edu.cn} \\
}

\begin{document}

\maketitle
\nolinenumbers
\begin{abstract}
Prospective memory means carrying out a deferred intention at the right future cue while other work continues.
Benchmarks now isolate it as an agent skill, yet frontier LLMs still struggle: the best published PM-Bench scaffold reaches only $65.1\%$ Set-F1.
We argue that this loop is schema-constrained state tracking rather than open-ended reasoning, and that small models can execute it when the action space is typed.
We propose the Prospective Intention Store (PIS) that puts lifecycle logic in code and scoped language work on the model.
The scaffold is agentic and training-free: no selector fine-tuning and no trajectory distillation.
On PM-Bench, DeepSeek-Chat with PIS reaches $82.9\%$ Set-F1.
On Gemma-E2B, Set-F1 is only $4.2\%$ without a store and at most $6.6\%$ under seven retrospective memories, while PIS reaches $66.2\%$.
PIS further reaches $70.1\%$ Set-F1, where retrospective memory methods stay at most $54.4\%$.
PIS sets a new state of the art on this benchmark and enables small models to surpass the published large-model scaffold.
\end{abstract}

\section{Introduction}
\label{sec:intro}
\vspace{-0.4em}

As language agents move from single-turn answering to long-horizon assistance---planning, tool use, and multi-session interaction---memory becomes indispensable~\citep{yao2023react,packer2023memgpt,sumers2023cognitive}.
Without an external store, models forget user constraints, lose intermediate state, and cannot reuse experience across steps~\citep{liu2024lost,maharana2024locomo,wu2025longmemeval}.
A large share of today's agent memory is retrospective: write notes from interaction, update or merge entries, and later retrieve what is relevant to a query~\citep{chhikara2025mem0,xu2025amem,zhong2024memorybank}.
In practice, that write--update--retrieve loop is dominated by \emph{narrow, repetitive} calls---structured extraction, short summaries, similarity lookup, light JSON edits---rather than open-ended reasoning~\citep{belcak2025slm,lightmem2025}.
Such scoped invocations are a natural fit for small language models (SLMs): they are cheaper, lower-latency, and deployable on-device, while large models remain a fallback for hard open-domain work~\citep{belcak2025slm,slm2025survey,erdogan2024tinyagent,chen2024octopusv2}.
Schema-constrained tool use already shows that 1B--7B models can match or exceed cloud-scale APIs when the action space is typed~\citep{erdogan2024tinyagent,chen2024octopusv2,patil2023gorilla}.

\begin{figure}[t]
\centering
\vspace{-1.3em}
\includegraphics[width=\linewidth]{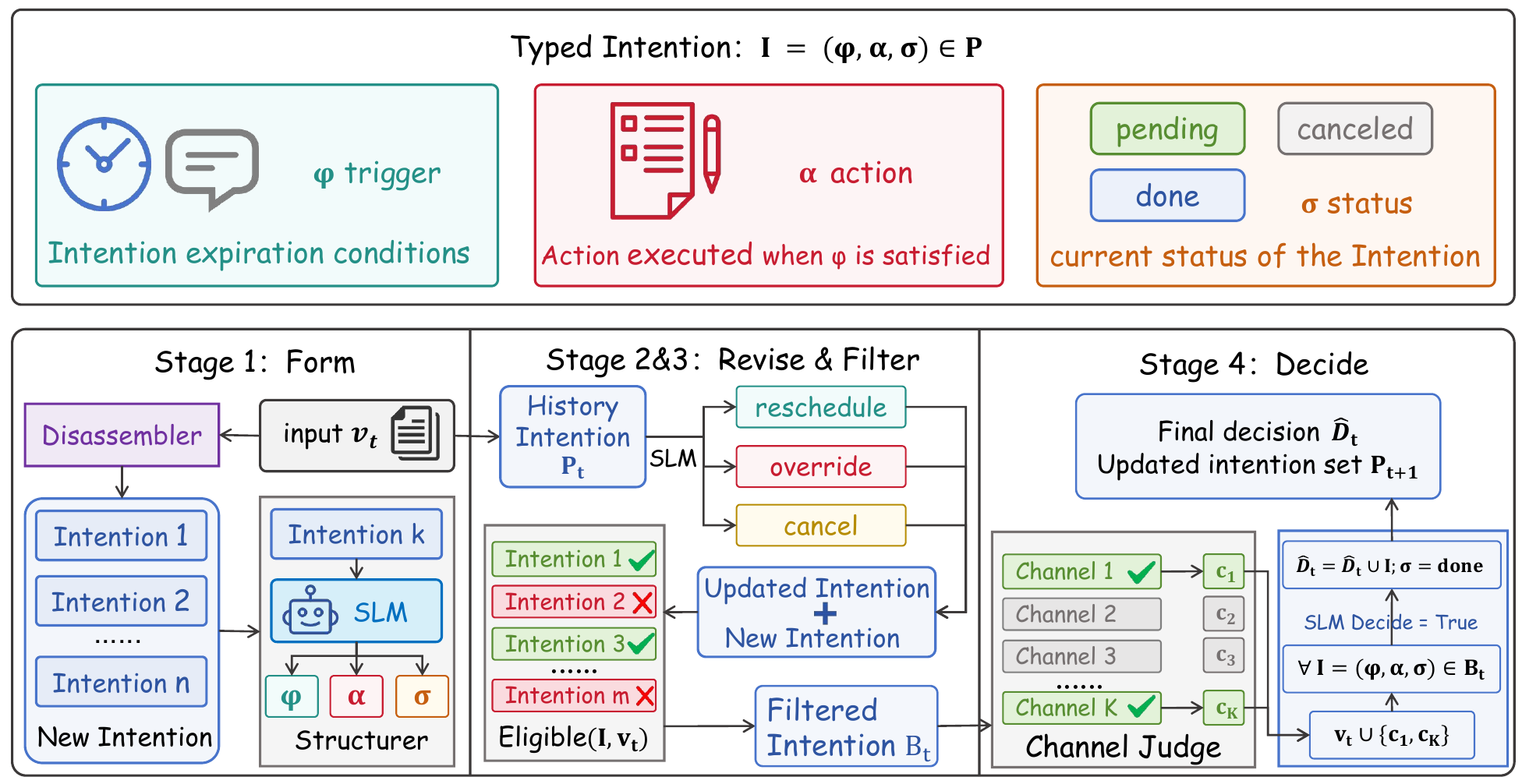}
\vspace{-1.0em}
\caption{Overview of the prospective intention store (PIS).
A deferred commitment is typed as $I{=}(\varphi,\alpha,\sigma)$; each step runs Form$\to$Revise$\to$Filter$\to$Decide (with optional channel enrichment before Decide) and emits a due set.}
\label{fig:overview}
\vspace{-0.3em}
\end{figure}

Recent Memory$\times$SLM systems report strong results: SLMs can run online controller/selector/writer pipelines~\citep{lightmemslm2026}, and dual-space distillation lets a 4B on-device agent approach a larger teacher when procedural memories are provided~\citep{duomem2026}.
Those gains typically require training-time cost---selector LoRA, teacher trajectories, or student adapters---before the small model operates memory well.
Despite this progress, the memories those systems operate remain retrospective: they retrieve or cache the past rather than maintain deferred commitments until a future cue fires.
Prospective memory (PM)---maintaining a deferred intention and executing it at a future cue or time while other activities continue---is ubiquitous in everyday assistance (``remind me when the portal opens,'' ``take the medicine at 21:00'') and has long been distinguished from retrospective recall in cognitive science~\citep{einstein1990pm,rendell2000virtualweek}.
In agent-memory work, PM remains sparsely studied: most benchmarks and stores optimize $\mathrm{sim}(e,q)$ rather than whether a trigger is satisfied and which set of actions is due~\citep{maharana2024locomo,wu2025longmemeval,xu2025amem,lightmem2025}.
Evaluations that isolate PM find that even frontier LLMs are unreliable---PM-Bench's best reported scaffold reaches only $65.1\%$ Set-F1~\citep{pmbench2026}, and TriggerBench shows near-ceiling retrospective accuracy where proactive intervention collapses~\citep{triggerbench2026}---and, to our knowledge, prior work has not studied SLMs for prospective memory.
Existing Memory$\times$SLM pipelines retrieve the past or cache procedural scripts; they do not maintain revisable due sets~\citep{duomem2026,lightmemslm2026}.
We argue that this gap is architectural as well as a scale gap: PM is schema-constrained state tracking of the same family as the repetitive memory operations SLMs already handle retrospectively~\citep{belcak2025slm,slm2025survey}.
\vspace{-0.2em}

Conditioned delayed memory requires three loops that similarity RAG does not provide: Update (rewrite cancellations and overrides), Decision (emit a due set), and Observation (expand clocks and channels into evidence)~\citep{pmbench2026,liu2024lost}.
Once intentions are typed as $(\varphi,\alpha,\sigma)$, those loops are schema-scoped---the regime where TinyAgent-style interfaces succeed~\citep{erdogan2024tinyagent}---but overload a small model if packed into one open JSON state machine.
We therefore propose the Prospective Intention Store (PIS; Figure~\ref{fig:overview}): an agentic scaffold that externalizes intentions as typed records, implements lifecycle and board logic in code, and leaves scoped grounding to the model, usable at inference without PEFT~\citep{belcak2025slm}.
On PM-Bench, PIS sets a new state of the art and lets small models exceed the published large-model scaffold.
\vspace{-0.3em}

\paragraph{Contributions.}
\begin{enumerate}
\item We provide, to our knowledge, the first systematic formalization of prospective-memory operations for language agents, casting deferred commitments as typed records and due-set decisions as operators over an external store.
\item We propose an agent system, PIS, that executes those operations as a training-free Form--Decide loop and show empirically that it yields strong PM-Bench results across large and small backbones.
\item We are the first to study SLMs for prospective memory: PIS not only surpasses large-model scaffolds in Set-F1, but also uses far fewer tokens and less wall-clock time than heavy large-model memory stacks, offering an efficient and low-cost solution for this skill~\citep{belcak2025slm}.
\end{enumerate}
\vspace{-0.2em}
We will release the method and code upon acceptance.

\vspace{-1em}
\section{Related Work}
\label{sec:related}
\vspace{-0.4em}

\paragraph{SLMs for scoped agent work.}
Belcak et~al.\ argue that most agent traffic is narrow and repetitive, so SLMs should be the default with selective LLM fallback~\citep{belcak2025slm,slm2025survey}.
TinyAgent and Octopus show 1B--7B models matching larger APIs when tool interfaces are typed~\citep{erdogan2024tinyagent,chen2024octopusv2}; we treat prospective memory as another such scoped subroutine.
\vspace{-0.5em}

\paragraph{Retrospective memory.}
MemGPT/Letta, Mem0, A-Mem, MemoryOS, LightMem and related stacks store and retrieve episodic or semantic notes~\citep{packer2023memgpt,chhikara2025mem0,xu2025amem,kang2025memoryos,lightmem2025}.
Memory$\times$SLM systems (LightMem-SLM, DuoMem) operate those pipelines with small models, typically after selector LoRA or teacher distillation~\citep{lightmemslm2026,duomem2026}.
Success metrics remain answer quality or environment reward rather than cue-triggered due-set Set-F1, and the stored object remains a note optimized for $\mathrm{sim}(e,q)$, not a revisable intention $(\varphi,\alpha,\sigma)$ with satisfaction $\mathrm{sat}(\varphi,V_t)$.
Overall, retrospective memory for agents is extensively studied, whereas work that targets prospective memory remains scarce.
\vspace{-0.5em}

\paragraph{Prospective memory.}
Cognitive work distinguishes event- and time-based deferred intentions~\citep{einstein1990pm,rendell2000virtualweek}.
PM-Bench isolates the skill for tool-using agents (best scaffold $65.1\%$ Set-F1); residuals concentrate on monitoring, cross-day obligations, and updates, consistent with missing structure rather than insufficient model scale~\citep{pmbench2026}.
TriggerBench shows prospective performance far below matched retrospective probes~\citep{triggerbench2026}.
PIS complements this landscape with a typed prospective object and Form--Decide operators on a frozen backbone, without selector LoRA or teacher distillation.

\section{Method}
\label{sec:method}
\vspace{-0.4em}

We cast conditioned delayed memory as operations over a typed prospective intention store (PIS; Figure~\ref{fig:overview}).
Lifecycle over fields is deterministic and structural; the language model performs only scoped semantic grounding.
This factorization is usable at inference without PEFT~\citep{belcak2025slm}.
\vspace{-0.25em}

\subsection{Problem formulation}
\label{sec:method-prelim}
\vspace{-0.25em}
At discrete step $t$, the agent observes evidence $V_t$ (vignette text, tool replies, clocks) and a menu $X_t$ of candidate actions.
It maintains an external store $\mathcal{P}_t$ of intentions.
Each intention is a triple $I=(\varphi,\alpha,\sigma)$, where $\varphi$ is the trigger condition, $\alpha$ is the intended action, and $\sigma\in\{\mathrm{pending},\mathrm{done},\mathrm{canceled}\}$ is status.
Beyond clocks and in-scene events, many real triggers depend on \emph{hidden state channels}---portals, dashboards, inboxes, sensors---whose contents are invisible until the agent proactively queries them.
We write $\mathcal{C}_t$ for the set of channels referenced by pending intentions in $\mathcal{P}_t$; a query on channel $c\in\mathcal{C}_t$ returns a reply that is merged into evidence, $V_t\!\leftarrow\!V_t\cup\{c\}$.
Without such Observation, $V_t\models\varphi$ cannot be evaluated for channel-conditioned pledges, so Decide must be allowed to request queries before emitting a due set.
Satisfaction $V_t\models\varphi$ is checked by rules for clocks and by a language judge for event and channel cues once evidence is present.
The ideal due set is $D^\star_t=\{\alpha\mid I\in\mathcal{P}_t,\ \sigma=\mathrm{pending},\ V_t\models\varphi\}$; the agent emits a predicted set $\hat{D}_t\subseteq X_t$, scored by trajectory Set-F1 against ground-truth $D_t$~\citep{pmbench2026}.
We write the decision and store update as $\hat{D}_t=\pi(\mathcal{P}_t,V_t)$ and $\mathcal{P}_{t+1}=F(\mathcal{P}_t,V_t,\hat{D}_t)$.
Operationally, each step factors into four operators---$\mathrm{Form}$, $\mathrm{Revise}$, $\mathrm{Filter}$, and $\mathrm{Decide}$---with an eligibility board $B_t\subseteq\mathcal{P}_t$ as the intermediate view on which Decide conditions, including which channels still need to be queried.
\vspace{-0.25em}

\subsection{Store and decision interface}
\label{sec:method-model}
\vspace{-0.25em}
The interface contracts are
\begin{align}
\hat{D}_t &= \pi(\mathcal{P}_t,V_t),
\mathcal{P}_{t+1} = F(\mathcal{P}_t,V_t,\hat{D}_t), \label{eq:decide&update}
\end{align}
\vspace{-0.35em}
with ideal due set
\begin{equation}
D^\star_t=\{\alpha \mid I\in\mathcal{P}_t,\ \sigma=\mathrm{pending},\ V_t\models\varphi\}.
\label{eq:dueset}
\end{equation}
Clock predicates are rule-checked; event and channel predicates use a language judge after any needed channel replies enter $V_t$.
\vspace{-0.25em}

\subsection{Operators}
\label{sec:method-ops}
\vspace{-0.25em}
Done-marking after execution is folded into Decide.

\paragraph{Form.}
Narrative is converted into typed rows once.
$\mathrm{Disassemble}(V_t)$ extracts commitment spans; $\mathrm{Structure}(s)=(\varphi,\alpha,\mathrm{pending})$ fills trigger, action, and status:
\begin{equation}
\Delta_t^{\mathrm{form}}=\mathrm{Form}(V_t)
=\{\mathrm{Structure}(s)\mid s\in\mathrm{Disassemble}(V_t)\},\quad
\mathcal{P}_t\leftarrow\mathcal{P}_t\cup\Delta_t^{\mathrm{form}}.
\label{eq:form}
\end{equation}
\vspace{-0.3em}
In the SLM realization, Structure is a slot-filling call assembled in code into $I{=}(\varphi,\alpha,\sigma)$; Disassemble and Structure may share one model invocation.
Form is the only stage where open text enters the store schema.
\vspace{-0.5em}
\paragraph{Revise.}
Revise performs belief update over the existing store rather than retrieval.
For each pending $I$, a language judge may apply one typed patch: \textbf{reschedule} rewrites $\varphi$, \textbf{override} replaces $\alpha$, or \textbf{cancel} sets $\sigma\!\leftarrow\!\mathrm{canceled}$.
Cue appearance alone is not cancellation; it is evidence for $V_t\models\varphi$ and is handled by Filter and Decide.
\begin{equation}
\rho_t(I;V_t)=
\begin{cases}
(\varphi',\alpha,\mathrm{pending}), & \mathrm{reschedule}(I;V_t),\\
(\varphi,\alpha',\mathrm{pending}), & \mathrm{override}(I;V_t),\\
(\varphi,\alpha,\mathrm{canceled}), & \mathrm{cancel}(I;V_t),\\
I, & \text{no revision},
\end{cases}
\quad
\mathrm{Revise}(\mathcal{P}_t,V_t)=\{\rho_t(I;V_t)\}_{I\in\mathcal{P}_t}.
\label{eq:revise}
\end{equation}
\vspace{-0.3em}
Implementation shortlists candidates $\mathcal{C}_t\subseteq\mathcal{P}_t$, then one indexed judge emits a sparse patch set applied in code (no update span implies an empty patch).
\vspace{-0.5em}
\paragraph{Filter.}
Pending rows are reduced to an eligibility board by structural rules (day/horizon, exact clock matches, discrete event/channel labels).
Channel-conditioned intentions whose watched channel has not yet been queried remain on the board as check targets rather than being dropped---a filtered view of $\mathcal{P}$, not embedding neighbors of $V_t$:
\begin{equation}
B_t=\mathrm{Filter}(\mathcal{P}_t,V_t)
=\{I\in\mathcal{P}_t \mid \sigma=\mathrm{pending},\ \mathrm{eligible}(I;V_t)\}.
\label{eq:filter}
\end{equation}
\vspace{-2.35em}

\paragraph{Decide.}
Because many pledges depend on hidden channels, a channel judge first inspects $B_t$ and may request one or more queries $c\in\mathcal{C}_t$; replies enrich evidence, $V_t\!\leftarrow\!V_t\cup\{c\}$, before due-set selection.
This Observation step mirrors everyday assistance, where the agent must open a portal or check a sensor rather than wait for the cue to appear in the dialogue.
Decide then maps $(B_t,V_t,X_t)$ to a due set and an updated store:
\begin{equation}
(\hat{D}_t,\mathcal{P}_{t+1})
=\mathrm{Decide}(B_t,V_t,X_t),
\label{eq:decide-op}
\end{equation}
\vspace{-0.3em}
where $\hat{D}_t\subseteq X_t$ and fulfilled pending rows are marked $\mathrm{done}$:
\begin{equation}
\kappa(I;\hat{D}_t)=
\begin{cases}
(\varphi,\alpha,\mathrm{done}), & \alpha\in\hat{D}_t,\ \sigma=\mathrm{pending},\ \text{guards allow},\\
I, & \text{otherwise,}
\end{cases}
\quad
\mathcal{P}_{t+1}=\{\kappa(I;\hat{D}_t)\mid I\in\mathcal{P}_t\}.
\label{eq:kappa}
\end{equation}
\vspace{-0.3em}
Filter and $\kappa$ are structural; Form, Revise, the channel judge, and Decide require language grounding~\citep{belcak2025slm}.
\vspace{-0.25em}

\subsection{One-step algorithm}
\label{sec:method-algo}
\vspace{-0.25em}
\begin{algorithm}[t]
\caption{One PIS step: realizing $(\pi,F)$}
\label{alg:pis-step}
\begin{algorithmic}[1]
\Require store $\mathcal{P}_{t}$, evidence $V_t$, menu $X_t$
\Ensure action set $\hat{D}_t$, updated store $\mathcal{P}_{t+1}$
\State $\mathcal{P}_{t}\leftarrow\mathcal{P}_{t}\cup\mathrm{Form}(V_t)$
\Comment{Eq.~\eqref{eq:form}}
\State $\mathcal{P}_{t}\leftarrow\mathrm{Revise}(\mathcal{P}_{t},V_t)$
\Comment{Eq.~\eqref{eq:revise}}
\State $B_{t}\leftarrow\mathrm{Filter}(\mathcal{P}_{t},V_t)$
\Comment{Eq.~\eqref{eq:filter}}
\While{channel judge on $B_t$ requests $c\in\mathcal{C}_t$}
\State $V_t\leftarrow V_t\cup\{c\}$
\Comment{query hidden channel}
\EndWhile
\State $(\hat{D}_t,\mathcal{P}_{t+1})\leftarrow\mathrm{Decide}(B_t,V_t,X_t)$
\Comment{Eqs.~\eqref{eq:decide-op}--\eqref{eq:kappa}}
\State \Return $\hat{D}_t,\ \mathcal{P}_{t+1}$
\end{algorithmic}
\end{algorithm}
\vspace{-0.6em}

Equivalently, $\pi$ comprises Filter, proactive channel Observation over $\mathcal{C}_t$, and Decide; $F$ comprises Form, Revise, and the $\kappa$ commit inside Decide.
Errors are attributable at operator boundaries (\S\ref{sec:exp}).

\section{Experiments}
\label{sec:exp}
\vspace{-0.4em}

\subsection{Setup}
\vspace{-0.25em}
We evaluate on the released PM-Bench synthetic week~\citep{pmbench2026}: seven simulated days of activity choice interleaved with anonymous prospective menus, lure actions, and hidden state channels.
The primary metric is micro \textbf{Set-F1} between predicted and ground-truth due sets; we also report update miss, cross-day miss, and false alarms per step (Table~\ref{tab:main}).
All backbones remain frozen: no selector LoRA, no teacher-trajectory distillation, and no PEFT, so gains are attributable to the memory scaffold.
Table~\ref{tab:main} compares training-free retrospective memories (Naive RAG; Mem0; A-Mem; Letta/MemGPT; LightMem-style; MemoryOS-style)~\citep{chhikara2025mem0,xu2025amem,packer2023memgpt,lightmem2025,kang2025memoryos} against PIS on DeepSeek-Chat~\citep{deepseek2024v3} and a local Gemma-E2B SLM~\citep{gemmateam2024gemma2}.
Table~\ref{tab:scale} uses the same layout on \texttt{Qwen3.5-4B} and \texttt{Qwen3-8B}~\citep{qwen3technicalreport}, with Gemma+PIS as a size reference.
Table~\ref{tab:compute} reports wall-clock duration and estimated main-prompt tokens.
LightMem/MemoryOS rows are pattern adapters of the corresponding retrieval designs rather than full upstream servers~\citep{lightmem2025,kang2025memoryos}.
The \texttt{single} row is a no-store baseline that reasons only from dialogue context.
\vspace{-0.25em}

\subsection{Main results}
\label{sec:main-results}
\vspace{-0.25em}

\paragraph{DeepSeek-Chat.}
PIS attains the highest Set-F1 at \textbf{$82.9\%$}, above every retrospective memory and the no-store single ($67.7\%$).
On the three secondary metrics it also leads: update miss drops to $22.2\%$ (vs.\ $55.6\%$--$100\%$ for the others), cross-day miss reaches $0.0\%$, and FA/step stays moderate at $8.8\%$---unlike Naive RAG or LightMem-style, which inflate false alarms with temporally obsolete notes.
Notably, single outperforms all retrospective memories ($46.5\%$--$58.3\%$).
Single relies on in-context reasoning over the ongoing dialogue rather than retrieved notes; the gap indicates that existing retrospective stores do not beat this PM-Bench baseline, whereas PIS does.

\paragraph{Gemma-E2B.}
Gemma-E2B is a short-context on-device-class SLM: when injected memory or long prompts exceed what it can use reliably, Set-F1 and the secondary metrics collapse.
All seven retrospective setups stay near floor ($0.0$--$6.6\%$ Set-F1), with update miss at least $88.9\%$, cross-day miss typically $100\%$, and often high FA/step under noisy retrieves.
We use this backbone deliberately to stress-test methods at the small-model extreme.
The same frozen PIS reaches \textbf{$66.2\%$}, about $10\times$ the best retrospective row, with cross-day miss falling to $28.6\%$ (update miss $77.8\%$ remains the main residual).

\begin{table*}[t]
\centering
\vspace{-0.8em}
\caption{PM-Bench main comparison (DeepSeek-Chat left, Gemma-E2B right).
\textbf{Set-F1}: micro $F_1$ over predicted vs.\ ground-truth due sets (higher is better).
\textbf{Update miss}: miss rate on the update-sensitive slice (reschedule/cancel/override; lower is better).
\textbf{Cross-day miss}: miss rate on intentions planted on an earlier day (lower is better).
\textbf{FA/step}: false alarms per decision step (lower is better).}
\label{tab:main}
\setlength{\tabcolsep}{2.0pt}
\footnotesize
\begin{tabular}{lcccc@{\hspace{0.9em}}cccc}
\toprule
& \multicolumn{4}{c}{DeepSeek-Chat} & \multicolumn{4}{c}{Gemma-E2B} \\
\cmidrule(lr){2-5}\cmidrule(lr){6-9}
Setup & Set-F1 & Update miss & Cross-day miss & FA/step
 & Set-F1 & Update miss & Cross-day miss & FA/step \\
\midrule
single
  & 67.7 & 55.6 & 57.1 & 6.3
  & 4.2 & 100.0 & 85.7 & 13.8 \\
+ Naive RAG
  & 46.5 & 88.9 & 100.0 & 57.5
  & 6.6 & 88.9 & 100.0 & 43.8 \\
+ Mem0
  & 48.7 & 100.0 & 57.1 & 6.3
  & 4.9 & 88.9 & 100.0 & 45.0 \\
+ A-Mem
  & 54.1 & 100.0 & 42.9 & 8.8
  & 0.0 & 100.0 & 100.0 & 10.0 \\
+ Letta
  & 52.2 & 66.7 & 71.4 & 25.0
  & 4.5 & 88.9 & 100.0 & 55.0 \\
+ LightMem-style
  & 51.5 & 66.7 & 57.1 & 38.8
  & 6.6 & 88.9 & 100.0 & 41.3 \\
+ MemoryOS-style
  & 58.3 & 66.7 & 42.9 & 8.8
  & 3.3 & 88.9 & 100.0 & 42.5 \\
+ PIS
  & \textbf{82.9} & 22.2 & 0.0 & 8.8
  & \textbf{66.2} & 77.8 & 28.6 & 15.0 \\
\bottomrule
\end{tabular}
\vspace{-1.0em}
\end{table*}

\subsection{Scaling to other SLMs}
\label{sec:scale}
\vspace{-0.25em}
Table~\ref{tab:scale} repeats the layout on \texttt{Qwen3.5-4B} and \texttt{Qwen3-8B}, with Gemma-E2B+PIS ($66.2\%$) as a size reference.
The two Qwen backbones tell a similar story: PIS is best on both ($70.1\%$ on 4B; $57.2\%$ on 8B), with lower update and cross-day misses than most retrieval rows, while retrospective memories again fail to provide a reliable gain over the no-store baseline (on 4B they all fall below single; on 8B a weak single leaves some retrieval rows above it, but none approach PIS).
Absolute scores are higher on \texttt{Qwen3.5-4B} than on \texttt{Qwen3-8B} because the 3.5-4B checkpoint is designed for agentic workloads, so it follows the typed board more reliably than the larger 8B chat model on this week.

\begin{table*}[t]
\centering
\vspace{-0.8em}
\caption{PM-Bench on other SLMs (same metrics and layout as Table~\ref{tab:main}).
Gemma-E2B+PIS is a size reference; columns are \texttt{Qwen3.5-4B} (left) and \texttt{Qwen3-8B} (right).}
\label{tab:scale}
\setlength{\tabcolsep}{2.0pt}
\footnotesize
\begin{tabular}{lcccc@{\hspace{0.9em}}cccc}
\toprule
& \multicolumn{4}{c}{Qwen3.5-4B} & \multicolumn{4}{c}{Qwen3-8B} \\
\cmidrule(lr){2-5}\cmidrule(lr){6-9}
Setup & Set-F1 & Update miss & Cross-day miss & FA/step
 & Set-F1 & Update miss & Cross-day miss & FA/step \\
\midrule
Gemma-E2B + PIS (ref.)
  & 66.2 & 77.8 & 28.6 & 15.0
  & 66.2 & 77.8 & 28.6 & 15.0 \\
\midrule
single
  & 57.4 & 88.9 & 85.7 & 18.8
  & 39.8 & 55.6 & 100.0 & 60.0 \\
+ Naive RAG
  & 48.8 & 88.9 & 85.7 & 57.5
  & 41.5 & 77.8 & 100.0 & 58.8 \\
+ Mem0
  & 54.4 & 88.9 & 100.0 & 50.0
  & 52.5 & 88.9 & 42.9 & 67.5 \\
+ A-Mem
  & 50.3 & 66.7 & 100.0 & 47.5
  & 45.7 & 55.6 & 71.4 & 45.0 \\
+ Letta
  & 43.9 & 44.4 & 100.0 & 86.3
  & 40.6 & 100.0 & 85.7 & 71.3 \\
+ LightMem-style
  & 46.9 & 66.7 & 100.0 & 86.3
  & 42.8 & 88.9 & 71.4 & 88.8 \\
+ MemoryOS-style
  & 50.3 & 77.8 & 85.7 & 67.5
  & 45.4 & 55.6 & 85.7 & 73.8 \\
+ PIS
  & \textbf{70.1} & 33.3 & 28.6 & 38.8
  & \textbf{57.2} & 55.6 & 57.1 & 47.5 \\
\bottomrule
\end{tabular}
\vspace{-1.0em}
\end{table*}

\subsection{Computation analysis}
\label{sec:compute}
\vspace{-0.25em}
SLM-agent deployments care about latency and token load as well as accuracy~\citep{belcak2025slm,slm2025survey}.
Table~\ref{tab:compute} reports wall-clock \textbf{Dur} and estimated main-prompt input tokens \textbf{Tok} (millions) from logged \texttt{EST\_INPUT\_TOKENS} on choose/query calls.
Side calls (intention judges, embedders, external memory servers) are excluded, so Tok is a lower bound; Dur includes side effects on the critical path.

\paragraph{Efficiency without competence.}
On Gemma and both Qwen SLMs, most retrospective scaffolds finish in a few minutes at most about $1$M estimated main tokens, yet remain near floor Set-F1.
That apparent efficiency largely reflects missing Observation traffic: few \texttt{check\_time}/\texttt{query\_state} loops shorten trajectories without improving PM.
A-Mem and Letta are expensive exceptions (tens of minutes; several million tokens) because linking and server-side LLM work dominate Dur.

\paragraph{PIS cost profile.}
PIS spends more on the main agent than a quiet single baseline (DeepSeek $16.4$\,min / $2.08$M; Gemma $4.7$\,min / $1.23$M; Qwen3.5-4B $6.5$\,min / $1.48$M; Qwen3-8B $5.6$\,min / $1.39$M) because channel enrichment and longer boards increase choose/query volume.
Relative to A-Mem and Letta, PIS is often cheaper in Dur while attaining the best Set-F1 on every backbone in Tables~\ref{tab:main}--\ref{tab:scale}.
For SLM stacks, a typed Form--Decide loop is preferable to heavy retrospective infrastructure that does not improve due-set F1.

\begin{table*}[t]
\centering

\caption{Computation on the PM-Bench week.
\textbf{Dur}: wall-clock time for one full week.
\textbf{Tok}: estimated main-prompt input tokens in millions (sum of logged \texttt{EST\_INPUT\_TOKENS}; excludes judge/embed side calls).}
\label{tab:compute}
\setlength{\tabcolsep}{2.2pt}
\footnotesize
\begin{tabular}{lrr@{\hspace{0.7em}}rr@{\hspace{0.7em}}rr@{\hspace{0.7em}}rr}
\toprule
& \multicolumn{2}{c}{DeepSeek} & \multicolumn{2}{c}{Gemma-E2B}
& \multicolumn{2}{c}{Qwen3.5-4B} & \multicolumn{2}{c}{Qwen3-8B} \\
\cmidrule(lr){2-3}\cmidrule(lr){4-5}\cmidrule(lr){6-7}\cmidrule(lr){8-9}
Setup & Dur & Tok & Dur & Tok & Dur & Tok & Dur & Tok \\
\midrule
single
  & 1.3m & 0.82 & 1.7m & 0.90
  & 1.7m & 0.73 & 2.1m & 1.04 \\
+ Naive RAG
  & 1.8m & 0.82 & 1.5m & 0.78
  & 2.3m & 0.72 & 3.3m & 0.99 \\
+ Mem0
  & 7.4m & 2.50 & 1.2m & 0.79
  & 2.1m & 0.75 & 2.5m & 0.87 \\
+ A-Mem
  & 22.0m & 7.84 & 18.4m & 3.37
  & 35.4m & 8.78 & 27.6m & 4.89 \\
+ Letta
  & 23.9m & 8.11 & 20.7m & 7.79
  & 14.8m & 5.23 & 15.2m & 6.23 \\
+ LightMem-style
  & 6.3m & 1.43 & 2.9m & 0.85
  & 3.5m & 0.75 & 5.5m & 1.42 \\
+ MemoryOS-style
  & 5.4m & 1.65 & 2.2m & 0.79
  & 3.5m & 0.72 & 6.1m & 1.47 \\
+ PIS
  & 16.4m & 2.08 & 4.7m & 1.23
  & 6.5m & 1.48 & 5.6m & 1.39 \\
\bottomrule
\end{tabular}
\vspace{-1.0em}
\end{table*}

\vspace{-1em}
\section{Conclusion}
\label{sec:conc}
\vspace{-0.4em}

Prospective memory is a newly isolated agent skill that frontier LLMs still struggle with, and that retrospective Memory$\times$SLM systems do not address.
We framed PM as schema-constrained state tracking and instantiated a typed intention store whose Form--Decide operators realize $(\pi,F)$ as agentic scaffolding without training a selector or distilling a student.
That training-free stance attributes competence to the typed object and the operator factorization rather than to PEFT as a prerequisite.
Large backbones with this store exceed published PM-Bench scaffolds ($82.9\%$ Set-F1 on DeepSeek-Chat); Gemma-E2B rises from at most $6.6\%$ under seven retrospective baselines to $66.2\%$ with the same frozen store; Qwen3.5-4B+PIS reaches $70.1\%$ while seven retrospective memories stay at most $54.4\%$.
Wall-clock and estimated token costs (\S\ref{sec:compute}) indicate that the quality gain comes from scaffolding rather than merely longer prompts.
\vspace{-0.25em}

\paragraph{Limitations.}
Public prospective-memory benchmarks for agents remain scarce; to our knowledge PM-Bench is the main openly available suite, so our evaluation is necessarily concentrated on this dataset.
Although PIS needs no model training, fine-tuning is not explored here and may further improve performance.
\vspace{-0.15em}

\paragraph{Future work.}
We plan to (i)~fine-tune models on the Form--Decide interfaces, (ii)~run evaluation studies that isolate the contribution of each operator, and (iii)~build additional prospective-memory benchmarks to complement the currently limited open supply beyond PM-Bench.
We release this as non-archival work in progress toward SLM-native prospective agents.

\small
\bibliography{references}

@article{belcak2025slm,
  title={Small Language Models are the Future of Agentic {AI}},
  author={Belcak, Peter and Heinrich, Greg and Diao, Shizhe and Fu, Yonggan and Dong, Xin and Muralidharan, Saurav and Lin, Yingyan Celine and Molchanov, Pavlo},
  journal={arXiv preprint arXiv:2506.02153},
  year={2025}
}

@article{erdogan2024tinyagent,
  title={{TinyAgent}: Function Calling at the Edge},
  author={Erdogan, Lutfi Eren and Lee, Nicholas and Jha, Siddharth and Kim, Sehoon and Tabrizi, Ryan and Moon, Suhong and Hooper, Coleman and Anumanchipalli, Gopala and Keutzer, Kurt and Gholami, Amir},
  journal={arXiv preprint arXiv:2409.00608},
  year={2024}
}

@article{chen2024octopusv2,
  title={Octopus v2: On-device language model for super agent},
  author={Chen, Wei and Li, Zhiyuan and Ma, Mingyuan},
  journal={arXiv preprint arXiv:2404.01744},
  year={2024}
}

@article{slm2025survey,
  title={Small Language Models for Agentic Systems: A Survey of Architectures, Capabilities, and Deployment Trade offs},
  author={Sharma, Raghav and Mehta, Manan},
  journal={arXiv preprint arXiv:2510.03847},
  year={2025}
}

@inproceedings{yao2023react,
  title={{ReAct}: Synergizing Reasoning and Acting in Language Models},
  author={Yao, Shunyu and Zhao, Jeffrey and Yu, Dian and Du, Nan and Shafran, Izhak and Narasimhan, Karthik and Cao, Yuan},
  booktitle={International Conference on Learning Representations},
  year={2023}
}

@article{patil2023gorilla,
  title={Gorilla: Large Language Model Connected with Massive {APIs}},
  author={Patil, Shishir G. and Zhang, Tianjun and Wang, Xin and Gonzalez, Joseph E.},
  journal={arXiv preprint arXiv:2305.15334},
  year={2023}
}

@article{xu2025amem,
  title={{A-Mem}: Agentic Memory for {LLM} Agents},
  author={Xu, Wujiang and Liang, Zujie and Mei, Kai and Gao, Hang and Tan, Juntao and Zhang, Yongfeng},
  journal={arXiv preprint arXiv:2502.12110},
  year={2025}
}

@article{chhikara2025mem0,
  title={{Mem0}: Building Production-Ready {AI} Agents with Scalable Long-Term Memory},
  author={Chhikara, Prateek and Khant, Dev and Aryan, Saket and Singh, Taranjeet and Yadav, Deshraj},
  journal={arXiv preprint arXiv:2504.19413},
  year={2025}
}

@article{packer2023memgpt,
  title={{MemGPT}: Towards {LLMs} as Operating Systems},
  author={Packer, Charles and Wooders, Sarah and Lin, Kevin and Fang, Vivian and Patil, Shishir G. and Stoica, Ion and Gonzalez, Joseph E.},
  journal={arXiv preprint arXiv:2310.08560},
  year={2023}
}

@article{kang2025memoryos,
  title={Memory {OS} of {AI} Agent},
  author={Kang, Jiazheng and Ji, Mingming and Zhao, Taize and Li, Fei},
  journal={arXiv preprint arXiv:2506.06326},
  year={2025}
}

@article{zhong2024memorybank,
  title={{MemoryBank}: Enhancing Large Language Models with Long-Term Memory},
  author={Zhong, Wanjun and Guo, Lianghong and Gao, Qiqi and Ye, He and Wang, Yanlin},
  journal={Proceedings of the AAAI Conference on Artificial Intelligence},
  volume={38},
  year={2024}
}

@article{duomem2026,
  title={{DuoMem}: Towards Capable On-Device Memory Agents via Dual-Space Distillation},
  author={Hosseini, Peyman and Bohdal, Ondrej and Alajrami, Ahmed and Maracani, Andrea and Castro, Ignacio and Purver, Matthew and Ozay, Mete and Ozkan, Savas and Ceritli, Taha},
  journal={arXiv preprint arXiv:2606.29961},
  year={2026}
}

@article{lightmem2025,
  title={{LightMem}: Lightweight and Efficient Memory-Augmented Generation},
  author={Fang, Jizhan and Deng, Xinle and Xu, Haoming and Jiang, Ziyan and Tang, Yuqi and Xu, Ziwen and Deng, Shumin and Yao, Yunzhi and Wang, Mengru and Qiao, Shuofei and Chen, Huajun and Zhang, Ningyu},
  journal={arXiv preprint arXiv:2510.18866},
  year={2025}
}

@article{lightmemslm2026,
  title={Lightweight {LLM} Agent Memory with Small Language Models},
  author={Zhang, Jiaquan and Zhang, Chaoning and Chen, Shuxu and Huang, Zhenzhen and Zheng, Pengcheng and Wang, Zhicheng and Guo, Ping and Mo, Fan and Bae, Sung-Ho and Zou, Jie and Wei, Jiwei and Yang, Yang},
  journal={arXiv preprint arXiv:2604.07798},
  year={2026}
}

@article{maharana2024locomo,
  title={Evaluating Very Long-Term Conversational Memory of {LLM} Agents},
  author={Maharana, Adyasha and Lee, Dong-Ho and Tulyakov, Sergey and Bansal, Mohit and Barbieri, Francesco and Fang, Yuwei},
  journal={arXiv preprint arXiv:2402.17753},
  year={2024}
}

@article{wu2025longmemeval,
  title={{LongMemEval}: Benchmarking Chat Assistants on Long-Term Interactive Memory},
  author={Wu, Di and Wang, Hongwei and Yu, Wenhao and Zhang, Yuwei and Chang, Kai-Wei and Yu, Dong},
  journal={arXiv preprint arXiv:2410.10813},
  year={2024}
}

@article{pmbench2026,
  title={{PM-Bench}: Evaluating Prospective Memory in {LLM} Agents},
  author={Liu, Genglin and Gabriel, Saadia},
  journal={arXiv preprint arXiv:2607.12385},
  year={2026}
}

@article{triggerbench2026,
  title={{TriggerBench}: Investigating Prospective Memory for Large Language Models},
  author={Zhang, Tianhua and Wang, Xinjiang and Zhang, Q. and Chen, Qi and Li, Kun and Chen, Yaoqi and Wang, Dingdong and Meng, Helen and Lu, Yan},
  journal={arXiv preprint arXiv:2606.23459},
  year={2026}
}

@article{einstein1990pm,
  title={Normal aging and prospective memory},
  author={Einstein, Gilles O. and McDaniel, Mark A.},
  journal={Journal of Experimental Psychology: Learning, Memory, and Cognition},
  volume={16},
  number={4},
  pages={717--726},
  year={1990}
}

@article{rendell2000virtualweek,
  title={Virtual week and actual week: Ageing and context dependent prospective memory},
  author={Rendell, Peter G. and Craik, Fergus I. M.},
  journal={Aging, Neuropsychology, and Cognition},
  volume={7},
  number={4},
  pages={209--226},
  year={2000}
}

@article{sumers2023cognitive,
  title={Cognitive Architectures for Language Agents},
  author={Sumers, Theodore and Yao, Shunyu and Narasimhan, Karthik and Griffiths, Thomas},
  journal={Transactions on Machine Learning Research},
  year={2024}
}

@article{liu2024lost,
  title={Lost in the Middle: How Language Models Use Long Contexts},
  author={Liu, Nelson F. and Lin, Kevin and Hewitt, John and Paranjape, Ashwin and Bevilacqua, Michele and Petroni, Fabio and Liang, Percy},
  journal={Transactions of the Association for Computational Linguistics},
  volume={12},
  pages={157--173},
  year={2024}
}

@article{deepseek2024v3,
  title={Deep{S}eek-{V}3 Technical Report},
  author={{DeepSeek-AI}},
  journal={arXiv preprint arXiv:2412.19437},
  year={2024}
}

@article{gemmateam2024gemma2,
  title={Gemma 2: Improving Open Language Models at a Practical Size},
  author={{Gemma Team} and others},
  journal={arXiv preprint arXiv:2408.00118},
  year={2024}
}

@article{qwen3technicalreport,
  title={Qwen3 Technical Report},
  author={{Qwen Team}},
  journal={arXiv preprint arXiv:2505.09388},
  year={2025}
}
\bibliographystyle{plainnat}

\end{document}